\documentclass{article} 
\usepackage{iclr2025_conference,times}

\usepackage{amsmath,amsfonts,bm}

\def\eqref#1{equation~\ref{#1}}

\def\1{\bm{1}}

\def\vg{{\bm{g}}}
\def\vh{{\bm{h}}}

\def\vx{{\bm{x}}}

\DeclareMathAlphabet{\mathsfit}{\encodingdefault}{\sfdefault}{m}{sl}
\SetMathAlphabet{\mathsfit}{bold}{\encodingdefault}{\sfdefault}{bx}{n}

\def\gD{{\mathcal{D}}}

\def\gX{{\mathcal{X}}}
\def\gY{{\mathcal{Y}}}

\def\sA{{\mathbb{A}}}

\def\sC{{\mathbb{C}}}

\def\sS{{\mathbb{S}}}

\def\sX{{\mathbb{X}}}
\def\sY{{\mathbb{Y}}}

\newcommand{\R}{\mathbb{R}}

\usepackage{hyperref}
\usepackage{url}

\usepackage{multirow}
\usepackage{paralist}
\usepackage{wrapfig}
\usepackage{graphicx}

\title{Generative Click-through Rate Prediction with Applications to Search Advertising\thanks{This work was first submitted on February 9, 2024.}}

\author{Lingwei Kong, Lu Wang, Changping Peng, Zhangang Lin, Ching Law \& Jingping Shao \\
JD.com\\
\texttt{\{konglingwei6,wanglu241,pengchangping\}@jd.com} \\
\texttt{\{linzhangang,lawching,shaojingping\}@jd.com} \\
}

\iclrfinalcopy 
\begin{document}

\maketitle

\begin{abstract}
Click-Through Rate (CTR) prediction models are integral to a myriad of industrial settings, such as personalized search advertising. Current methods typically involve feature extraction from users' historical behavior sequences combined with product information, feeding into a discriminative model that is trained on user feedback to estimate CTR. With the success of models such as GPT, the potential for generative models to enrich expressive power beyond discriminative models has become apparent. In light of this, we introduce a novel model that leverages generative models to enhance the precision of CTR predictions in discriminative models. To reconcile the disparate data aggregation needs of both model types, we design a two-stage training process: 1) Generative pre-training for next-item prediction with the given item category in user behavior sequences; 2) Fine-tuning the well-trained generative model within a discriminative CTR prediction framework. Our method's efficacy is substantiated through extensive experiments on a new dataset, and its significant utility is further corroborated by online A/B testing results. Currently, the model is deployed on one of the world's largest e-commerce platforms, and we intend to release the associated code and dataset in the future.
\end{abstract}

\section{Introduction}

Click-through rate (CTR) prediction is essential in online advertising, particularly within the cost-per-click (CPC) revenue model where advertisers pay for each click on their sponsored items.  CTR prediction models analyze historical ad impressions and click logs to estimate the probability a user will click on a given item. These estimations are then used alongside advertising bids to determine which ads to display, impacting user experience and advertiser ROI. Traditionally, CTR prediction models leaned on discriminative methods, using features from user interactions and product details to forecast responses~\citep{zhou2018din,song2021underestimation,zheng2022implicit,wu2021survey,wang2023pluggable,kong2023lovf}.

We propose a groundbreaking approach, GenCTR (Generative CTR prediction), that leverages the potential of generative models to enhance the representational power of click-through rate (CTR) prediction systems. Our model aims to surpass the predictive accuracy of conventional discriminative models by effectively capturing intricate patterns within user behavior. GenCTR employs a unique dual-phase training methodology to reconcile the distinct data processing needs of generative and discriminative approaches. In the initial phase, generative pre-training establishes a robust understanding of user preferences by predicting the subsequent item within a user's interaction sequence. This pre-trained model is then fine-tuned within a discriminative framework optimized for CTR prediction in the second phase.

To effectively harness the inherent advantages of GenCTR, we introduce four core techniques: conditional self-condition decoder and conditional negative sampling in the generative pre-training stage, followed by parameter sharing and model integration in the discriminative fine-tuning stage.

We demonstrate GenCTR's practical value through its successful deployment in a large-scale online search advertising system within one of the world's leading e-commerce platforms, serving hundreds of millions of users daily. To foster continued research advancements, we are releasing a novel public dataset collected from real-world traffic. This dataset includes both pre-training and fine-tuning data, providing a valuable resource for the research community.

\section{Related Work}
\subsection{Click-Through Rate Prediction Models}
The pursuit of accurate click-through rate (CTR) prediction has led to the development of numerous models, which serve as a cornerstone for personalization in online advertising and recommendation systems. Early models, such as logistic regression, were limited by their linearity and inability to capture complex feature interactions. The introduction of the Wide \& Deep model by Cheng et al.~\citep{WideDeep} represented a significant advance, combining the memorization capabilities of wide linear models with the generalization abilities of deep neural networks.
Following this, DeepFM~\citep{DeepFM} and xDeepFM~\citep{xDeepFM} models further enhanced the feature interaction modeling by integrating factorization machines into the deep learning framework. These models, however, largely focused on static user features without considering the temporal dynamics of user behavior.
To address the static nature of user representation in traditional CTR models, researchers began exploring models that could incorporate dynamic user behaviors. The Deep Interest Network (DIN) proposed by Zhou et al.~\citep{DIN} was a pioneering work in this area, introducing an attention mechanism to weigh user's historical behaviors differently when predicting CTR for different items. This model was able to capture the varying relevance of past interactions with respect to the current ad or item.Subsequently, researchers have explored two-stage approaches such as SIM~\citep{SIM}, ETA~\citep{ETA} and TWIN~\citep{TWIN}, which have successfully incorporated a broader spectrum of user historical behaviors while ensuring manageable serving complexity. These methodologies have yielded promising results in enhancing the accuracy of user interest predictions.


While these advancements have significantly improved the utilization of user historical data, there remains untapped potential in exploiting the rich positive feedback from users. In our work, we take a step further by pre-training generative models on user behaviors, leveraging user positive feedback as labels.

\subsection{Sequential Recommendation}
Sequential recommendation systems aim to predict the next item a user is likely to interact with by considering the order of past interactions. The evolution of these models began with the application of Convolutional Neural Networks (CNNs) and Recurrent Neural Networks (RNNs) to capture the sequential patterns in user-item interactions. CNN-based models, such as Caser~\citep{caser}, applied convolutional filters to learn local features of interaction sequences. RNN-based approaches, including GRU4Rec~\citep{GRU4Rec}, utilized gated recurrent units to model the user's sequential behavior dynamically.
The introduction of attention mechanisms marked a significant advancement in sequential recommendation models. The Self-Attentive Sequential Recommendation (SASRec) model by Kang et al.~\citep{SASRec} leveraged self-attention to identify which past items are most predictive of future interactions. This model was able to outperform traditional RNN-based models by focusing on the most relevant parts of a user's interaction history. Building on the success of attention-based models, Bert4Rec by Sun et al.~\citep{Bert4Rec} incorporated the bidirectional encoder representations from transformers (BERT) architecture to model user sequences. Bert4Rec's bidirectional training strategy provided a more comprehensive understanding of the context surrounding each item in the sequence, leading to improved recommendation performance.

While the aforementioned models have significantly advanced the field of sequential recommendation, our work diverges by focusing on the application of sequential recommendation techniques in the context of CTR prediction. We utilize sequential recommendation models as a pre-training step to enrich the feature space of our CTR prediction model. This pre-training step allows our model to capture the temporal dynamics of user behavior, which is often overlooked in traditional CTR models.

\section{Preliminaries}

\begin{figure}
    \centering
    \includegraphics[width=\columnwidth]{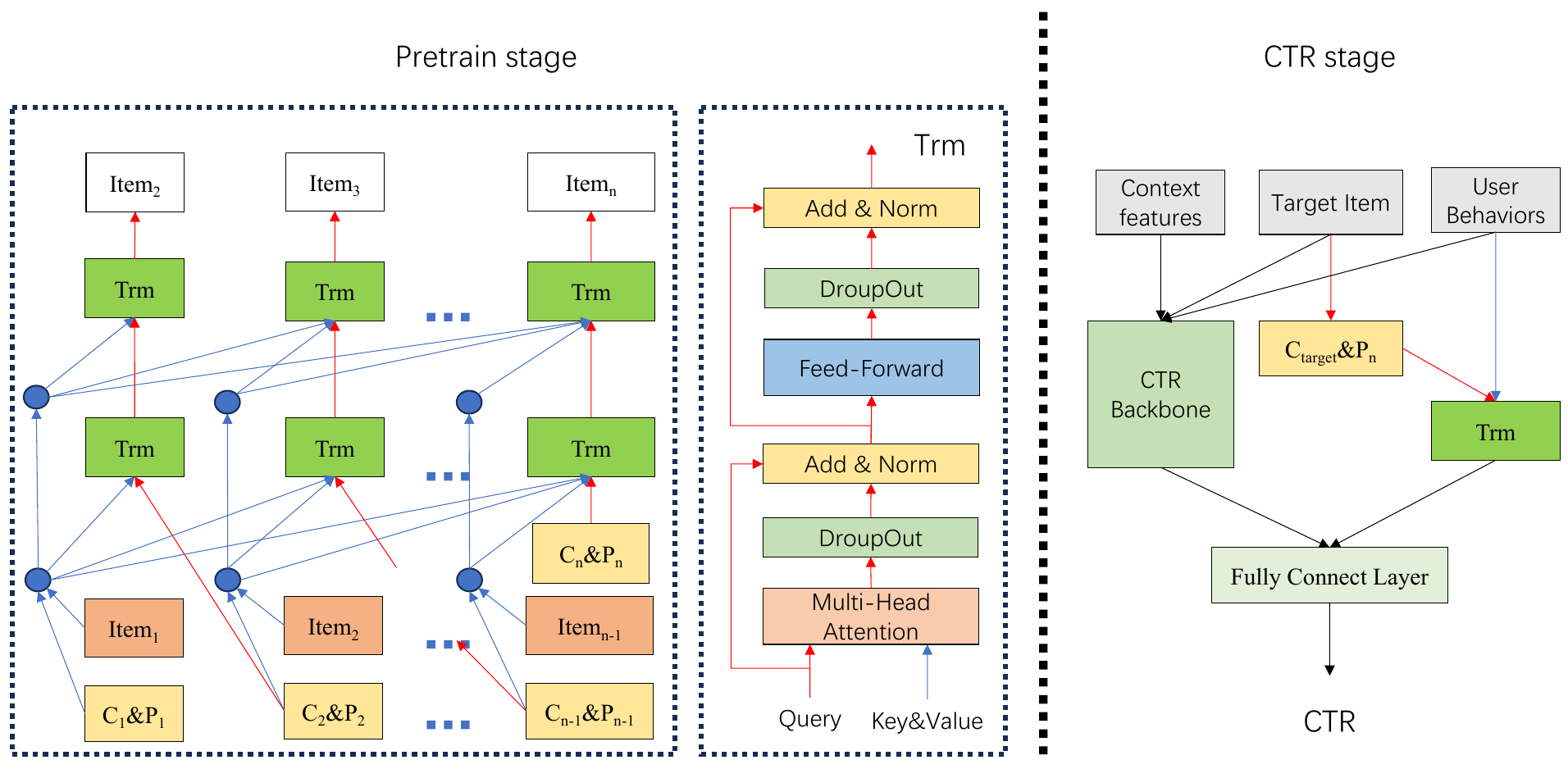}
    \caption{Overall architecture of GenCTR}
    \label{fig:arch}
\end{figure}

We denote the input space by $\gX$, encompassing all features associated with the user, the item, and the relevant context. The output space, $\gY = \{0, 1\}$, signifies whether a click occurs. Let $\gD$ denote the underlying distribution over $\gX \times \gY$.
Click-through rate (CTR) prediction seeks to develop a model, $\hat{y}: \gX \rightarrow (0, 1)$.
This model estimates the probability of a click happening (i.e., the CTR) for any given input $\vx \in \gX$,
with $\hat{y}(x)$ essentially predicting $\Pr_{y \sim \gD_{\sY \mid x}}\{y=1 \mid x\}$.

Each user can possess a unique user behavior sequence,
representing a chronological record of their interactions with items (e.g., clicks, purchases and adding items to cart).
This sequence can be further enriched with side information attached
to each item, providing deeper context.
Let $\sA$ be the space of items and $\sC$
be the space of side information corresponding to these items.
Then, the sequence space of user behavior is denoted by $(\sA \times \sC)^*$.
By incorporating this sequence as an auxiliary user feature,
CTR prediction models can leverage valuable insights
into user preferences and historical engagement,
ultimately enhancing prediction accuracy.


\section{Proposed method}

We introduce GenCTR, a novel two-stage generative method for enhanced click-through rate (CTR) prediction. GenCTR augments traditional backbone CTR models by leveraging user behavior sequences through pre-training.

\subsection{Pre-training}

The first step is pre-training the generative model.

\paragraph{Generative model}
Given a user's behavior sequence, or a continuous segment within it,
a generative model aims to predict the next item in the sequence by generating an embedding that captures its potential characteristics.
This embedding is a numerical representation of the item, typically denoted by $\vh \in \R^d$,
where $d$ represents the embedding dimension.

\paragraph{Conditional generative model}
Instead of solely relying on the historical sequence of items,
we propose enhancing the behavior sequence utilization by augmenting the generative model to consider side information associated with the next item.
This side information acts as an additional input, alongside the previous items, when predicting the next element.
Essentially, our model becomes conditionally dependent on this incoming side information.
Mathematically, we can express the generative model as a function $\vg: (\sA \times \sC)^* \times \sC \rightarrow \R^d$.

\paragraph{Conditional self-attention decoder}
Our conditional generative model is implemented as a variant of the self-attention decoder.
In a conventional self-attention decoder, the query vector is derived from the most recent item,
while the key and value vectors are derived from items that precede it.
In contrast, our conditional self-attention decoder adopts a distinct approach
--- its query vector is derived from the side information associated with the next item,
yet the key and value vectors come from the preceding items (including the most recent one) for the next item.
We term the method as the \emph{conditioned self-attention decoder}.
Figure~\ref{fig:arch} (left) illustrates the generative model's overall architecture. Our conditional self-attention decoder employs a decoder-only multi-layered multi-head transformer architecture. Notably, Position embedding is a 16 dimensional embedding with the length of user behavior.

\paragraph{Conditional negative sampling}
Negative sampling plays an important role in training self-attention decoders to improves model discrimination.
Corresponding to the conditional self-attention mechanism,
negative instances are sampled from the items which share the same side information of the next item.

\paragraph{Loss function}
Consider a user behavior sequence.
For each item, the decoder masks the item (preserving the side information used for the query vector)
and its subsequent items.
It then predicts an embedding of the current item.
The true label is simply the item itself,
while negative labels are sampled using conditional negative sampling.
Cross-entropy loss is then calculated for each prediction.
This process results in a sequence of losses, with the total number matching the sequence length.

\subsection{Integration}

This step seamlessly merges the generative model
with the backbone CTR prediction model.
Two key integration mechanisms are employed:

\paragraph{Parameter sharing}
Both models often share features like item identifiers and their side information.
This common ground is leveraged by utilizing the embedding parameters
learned by the generative model for these shared features
within the backbone CTR prediction model.
This effectively allows the backbone model
to benefit from the insights captured by the generative model
for these specific features.

\paragraph{Model integration}
Beyond parameter sharing, the models are directly connected,
resulting in the integrated CTR prediction model.
The output of the generative model
is fed as an additional input to the backbone CTR prediction model.
This enriches the feature space with the latent context
and potential interactions identified by the generative model.
Specifically, given the input tuple $(x, s, c)$,
where $x \in \sX$ is the original features for the backbone CTR model,
$s \in (\sA \times \sC)^*$ is the user behavior sequence,
$c$ is the side information for the target item,
the output of the integrated CTR model
is $\hat{y}(x, g(s, c))$,
where $g(s, c)$ denotes the output of the generative model.

The integrated model is trained end-to-end on the training data 
$\sS = \{(x, s, c, y)\}$,
solving the following optimization problem:
\begin{align}
    \min_\theta \frac{1}{\lvert\sS\rvert}
    \sum_{(x, s, c, y) \in \sS} \ell(y, \hat{y}(x, g(s, c; \theta)); \theta),
\end{align}
where $\ell(\cdot, \cdot)$ is the cross entropy loss,
$\theta$ encompasses all parameters of the backbone CTR prediction model
and the generative model.

Note that the backbone model already utilizes information from $s$ and $c$.
The generative model further enriches this representation
by capturing potential interactions and latent patterns
within the sequence and contextual data.

\section{Experiments}

\begin{table*}[!ht]
\caption{Experimental results on GCTR. Each experiment is repeated for 5 times independently (mean$\pm$std). $\uparrow$ means the larger number is better; $\downarrow$ means the smaller number is better.}
\label{tab:mainresults}
    \centering
    \resizebox{1.0\linewidth}{!}{
    \begin{tabular}{c|c|c|c|c|c|c|c|c|c}
    \hline
        ~ & \multicolumn{3}{c|}{BackBone} & \multicolumn{2}{c|}{DNN} & \multicolumn{2}{c|}{DCN V2} & \multicolumn{2}{c}{DCN V2\&TA} \\ \hline
No. & CTR Model & Integration Setting & Pre-train Setting & AUC$\uparrow$ & Logloss$\downarrow$ & AUC$\uparrow$ & Logloss$\downarrow$ & AUC$\uparrow$ & Logloss$\downarrow$  \\ \hline
        1 & Backbone &  &  & 0.6072±0.0002 & 0.2381±0.0000 & 0.6088±0.0000 & 0.2397±0.0000 & 0.6136±0.0001 & 0.2373±0.0000  \\ \hline
        2 & Backbone & PS & CS+CD & 0.6258±0.0003 & 0.2373±0.0001 & 0.6161±0.0013 & 0.3083±0.0082 & 0.6292±0.0006 & 0.2363±0.0000  \\ \hline
        3 & +decorder &  &  & 0.6124±0.0003 & 0.2375±0.0001 & 0.6141±0.0002 & 0.2373±0.0000 & 0.6150±0.0002 & 0.2372±0.0000  \\ \hline
        4 & +decorder & PS & CS+CD & 0.6310±0.0004 & 0.2371±0.0002 & 0.6292±0.0003 & 0.2371±0.0002 & 0.6299±0.0008 & 0.2362±0.0001  \\ \hline
        5 & \textbf{+decorder} & \textbf{PS+MI} & \textbf{CS+CD} & \textbf{0.6331±0.0007} & \textbf{0.2368±0.0002} & \textbf{0.6329±0.0003} & \textbf{0.2368±0.0002} & \textbf{0.6323±0.0007} & \textbf{0.2361±0.0000}  \\ \hline
        6 & +decorder & PS+MI & CS+SD & 0.6328±0.0004 & 0.2368±0.0001 & 0.6304±0.0004 & 0.2369±0.0001 & 0.6318±0.0003 & 0.2361±0.0000  \\ \hline
        7 & +decorder & PS+MI & RS+CD & 0.6259±0.0003 & 0.2363±0.0000 & 0.6258±0.0002 & 0.2362±0.0000 & 0.6263±0.0016 & 0.2364±0.0002  \\ \hline
        8 & +decorder & PS+MI & RS+SD & 0.6155±0.0006 & 0.2374±0.0001 & 0.6123±0.0002 & 0.2376±0.0000 & 0.6142±0.0004 & 0.2377±0.0000 \\ \hline
    \end{tabular}
    }
\end{table*}
Since there is no public dataset available containing both search query and user behavior information
for CTR prediction
(to the best of our knowledge),
 we collect a new dataset named GCTR from the online search advertising logs of one of the world’s largest e-commerce platforms.
 The GCTR dataset is structured into four distinct parts: the training set, the test set, the pre-training set, and the training sampling table. The training set is constructed from a sampling of logs over a span of three consecutive days and the test set is derived from the logs of the subsequent day.
The dataset encompasses detailed item information, user profiles, and users' last 200 clicks on items with corresponding side info. 

In addition to the primary training and test sets, we have also curated a pre-training set and a category-item sampling table. The pre-training set is generated by deduplicating user entries from the training set, highlighting user profiles and previous browsing activities. The category-item sampling table, on the other hand, records a mapping of all product categories to their corresponding items as they appear in the training set. 
Our subsequent offline experiments are conducted on this dataset, providing a robust foundation for validating the effectiveness of various CTR models.

\subsection{Evaluation metrics}



Two metrics are employed to assess the offline performance: AUC and LogLoss. AUC, or Area Under the Receiver Operating Characteristic Curve, is the paramount offline metric for Click-Through Rate (CTR) prediction, gauging the overall ranking accuracy. LogLoss, representing the model's cross-entropy loss, reflects its classification efficacy. Additionally, for online business impact, we measure CTR and RPM (Revenue Per Thousand Impressions) to validate our method's practical utility in an actual advertising ecosystem.

\subsection{Experimental Setup}

To evaluate the effectiveness of our proposed approach, we conduct experiments using three distinct backbone models for CTR prediction:

\begin{inparaenum}
    \item \textbf{DNN}: A basic neural architecture comprising an embedding layer followed by a multi-layer perceptron (MLP) with dimensions 256 and 128, utilizing ReLU activation functions. 
    
    \item \textbf{DCN V2~\citep{DCN_v2}}: The model is the state-of-the-art (SOTA) for CTR prediction without incorporating user behavior. We set the model with 3 experts, a cross-layer depth of 3, and a low-rank cross layer with a rank of 16.
    
    \item \textbf{DCN V2 \& TA}: Building on DCN v2, we integrate Target Attention (TA) to model user behavior, leveraging its proven efficacy in the state-of-the-art Exact Search Unit (ESU) for long-term user behavior analysis~\citep{SIM,TWIN,ETA}. This fusion aims to validate our method's impact on CTR models already incorporating user behavior.
\end{inparaenum}

All models are trained on NVIDIA P40 GPUs using TensorFlow. We maintain uniformity in our experimental setup by employing a single-layer decoder, setting the embedding dimension to 16 for all features, and using the Adam optimizer with a learning rate of 0.001. User behavior sequences are truncated to the last 200 interactions. Models undergo pre-training for three epochs and CTR training for one epoch. Our experimental findings are summarized in Table~\ref{tab:mainresults} with configurations denoted as follows:

\begin{inparaenum}
    \item \textbf{+decoder}: Introduce the pre-train model structure and fuse it with the backbone before the final discriminator layer.
    \item \textbf{PS}: Parameter sharing with pre-trained embeddings.
    \item \textbf{MI}: Inheritance of pre-trained model structure and parameters.
    \item \textbf{CS}: Category-Conditioned Negative Sampling in pre-training.
    \item \textbf{RS}: Random negative sampling in pre-training.
    \item \textbf{CD}: Category-Conditional decoder in pre-training.
    \item \textbf{SD}: Self-attention decoder similar to SASRec without conditioning.
\end{inparaenum}

Our setup is designed to critically assess the impact of various model components and training strategies on the CTR prediction task.


\subsection{Experiment Analysis}

\paragraph{Effect of Parameter Inheritance} 
Experiments 1, 2, 3, and 4 were designed to evaluate the impact of sharing pre-trained embedding parameters on CTR models. The significant improvements observed across all CTR models, without introducing any additional parameters to the CTR models.  This confirms that the item representations refined through our pre-training process have a pronounced positive effect on CTR estimation.

\paragraph{Effect of Model Inheritance}
The analysis contrasting experiments 2 and 5 underscores the distinct advantages of model inheritance. By comparing experiments 4 and 5, we controlled for model complexity and confirmed that the performance gains are not merely due to an expanded parameter count, but rather to the strategic advantage of model inheritance.

Building upon the significant gains from embedding and intermediate parameter incorporation, we delved into the influence of the pre-training model's design on CTR prediction. Experiment 5, representing our ultimate model configuration, served as the benchmark for subsequent ablation studies aimed at elucidating this aspect.

\paragraph{Effect of Pre-training with Negative Sampling Methods}
Comparisons between experiments 5 and 7, as well as 6 and 8, reveal that Category-Conditioned Negative Sampling (CS) consistently outperforms Random Negative Sampling (RS) in different pre-training contexts. The enhanced performance is attributed to CS's nuanced focus on item differences within categories, which is crucial for refining representations and improving CTR prediction accuracy.

\paragraph{Effectiveness of Pre-training with Decoder Modeling}
The performance contrast between experiments 5, 6 and 7, 8 reveals that our Conditional Decoder (CD) outshines the Self-attention Decoder (SD) across all backbone models. We posit that the CD's focus on item-specific features—as opposed to category-level characteristics accessible during CTR prediction—aligns the pre-training objectives more closely with the CTR modeling task. This alignment ensures that the pre-training phase concentrates on the intrinsic properties of items, thereby making the pre-trained model more relevant and effective for CTR prediction.

\paragraph{Online A/B testing}
We evaluated GenCTR through a two-week online A/B test on our live search advertising platform in October 2023.
This test served to evaluate the model's performance under real-world traffic conditions.
Notably, even the introduction of pre-trained embeddings through parameter sharing (PS) alone led to a significant uplift of $0.56\%$ in Click-Through Rate (CTR) and $1.25\%$ in Revenue Per Mille (RPM).
It was further enhanced upon integrating the full pre-trained model via model inheritance (MI), which resulted in an additional increase of $1.32\%$ in CTR and $1.66\%$ in RPM. These improvements substantiate the effectiveness of both inheritance strategies.

It is important to highlight that the baseline model used for comparison was the previously serving model on the real traffic, which already exhibited superior performance. Our model has now been successfully integrated and is currently serving the search advertising, further demonstrating the practical value of our research.

\section{Conclusion}

In this paper, we have presented a novel strategy that employs pre-trained generative models to harness user behavior data for enhancing CTR prediction models. Our approach is characterized by a bespoke design of the pre-training phase, aimed at extracting and utilizing the intricate patterns of user interactions to inform CTR predictions.
The empirical validation of our method through extensive offline experiments has confirmed its effectiveness, with a marked improvement in prediction accuracy. Additionally, its successful deployment in a search advertising system demonstrates the model's practical applicability.
In summary, our work opens avenues for future research to explore the full potential of generative pre-training in CTR modeling.

\bibliography{reference}
\bibliographystyle{iclr2025_conference}

\appendix

\end{document}